\documentclass[runningheads]{llncs}
\usepackage[T1]{fontenc}
\usepackage{graphicx,verbatim}
\usepackage{xcolor}
\usepackage{booktabs}
\usepackage{multirow}
\usepackage{amsmath}
\usepackage{soul}
\usepackage{amssymb}
\usepackage{siunitx}
\usepackage{makecell}
\usepackage{tabularx}
\usepackage{bbm}

\usepackage[colorlinks=true,
            linkcolor=blue,
            urlcolor=red,
            citecolor=green,
            pdfborder={0 0 0},
            pdfhighlight=/I]{hyperref}
\usepackage{xurl}     
\hypersetup{
  pdftitle={DUCX: Decomposing Unfairness in Tool-Using Medical Chest X-ray Agents},
  pdfauthor={Zikang Xu, Ruinan Jin, Xiaoxiao Li},
  pdfsubject={Fairness in medical agents and tool-using chest X-ray AI agents},
  pdfkeywords={fairness in medical agents, fairness in agents, medical AI agents, clinical agents, LLM agents, chest X-ray agents, algorithmic fairness, healthcare AI fairness, agentic medical imaging}
}

\newcommand{\ours}{\texttt{DUCX}}

\newcommand{\xbest}[1]{\textcolor{red}{#1}}
\newcommand{\xsecond}[1]{\textcolor{blue}{#1}}

\begin{document}
\title{\texorpdfstring{\includegraphics[width=0.04\linewidth]{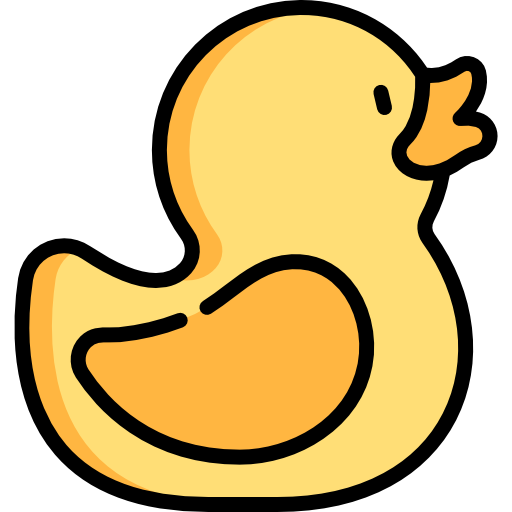}\ours: Decomposing Unfairness in Tool-Using Medical Chest X-ray Agents}{DUCX: Decomposing Unfairness in Tool-Using Medical Chest X-ray Agents}}
\titlerunning{Fairness in Medical Chest X-ray Agents}
%

\author{Zikang Xu\inst{1}$^*$ \and 
Ruinan Jin\inst{2}\inst{3}$^*$ \and
Xiaoxiao Li\inst{2}\inst{3}}  


\authorrunning{Xu, Z. et al.}
\institute{\textsuperscript{1} Institute of Artificial Intelligence, Hefei Comprehensive National Science Center, Anhui, China\\ \textsuperscript{2} The University of British Columbia, Vancouver, BC V6Z 1Z4, Canada\\
\textsuperscript{3} Vector Institute, Toronto, ON M5G 1M1, Canada \\
    \email{zikangxu@mail.ustc.edu.cn, ruinanjin@alumni.ubc.ca, xiaoxiao.li@ece.ubc.ca}}
  
\maketitle              
\begin{abstract}
Fairness in medical agents is becoming critical as tool-using clinical AI systems orchestrate specialized vision and language modules for tasks such as chest X-ray question answering. While these medical AI agents can improve flexibility, their added pipeline complexity also creates new pathways for demographic bias beyond standalone models. We present \ours{} (\textbf{D}ecomposing \textbf{U}nfairness in \textbf{C}hest \textbf{X}-ray agents), a systematic audit of fairness in tool-using chest X-ray agents instantiated with MedRAX. To localize where disparities arise, we introduce a stage-wise fairness decomposition that separates \emph{end-to-end bias} from three agent-specific sources: \emph{tool exposure bias} (utility gaps conditioned on tool presence), \emph{tool transition bias} (subgroup differences in tool-routing patterns), and \emph{model reasoning bias} (subgroup differences in synthesis behaviors). Extensive experiments on tool-used based agentic frameworks across five driver backbones reveal that (i) demographic gaps persist in end-to-end performance, with equalized odds up to 20.79\%, and the lowest fairness-utility tradeoff down to 28.65\%, and (ii) intermediate behaviors, tool usage, transition patterns, and reasoning traces exhibit distinct subgroup disparities that are not predictable from end-to-end evaluation alone (e.g., conditioned on segmentation-tool availability, the subgroup utility gap reaches as high as 50\%). Our findings underscore the need for process-level fairness auditing and debiasing to ensure the equitable deployment of clinical agentic systems. Code is available at \url{https://github.com/Nanboy-Ronan/DUCK}.


\end{abstract}
{%
  \let\thefootnote\relax%
  \footnotetext{$^*$Ruinan Jin and Zikang Xu contributed equally to this paper. \\ Corresponding author: Xiaoxiao Li}%
  \let\thefootnote\svthefootnote%
}

\section{Introduction}

Artificial intelligence (AI) is increasingly integrated into medical imaging, supporting tasks from disease detection to clinical decision assistance~\cite{zhou2021review}. Despite strong performance gains from vision and vision-language models, growing evidence shows that medical AI can exhibit demographic disparities across patient subgroups (e.g., age and gender), raising concerns about safety, reliability, and equitable deployment~\cite{xu2024addressing,liu2023translational}.

Most fairness studies in medical AI have focused on \ul{\emph{standalone}} models~\cite{xu2024addressing,jin2024debiased}, including convolutional networks~\cite{xu2023fairadabn}, vision foundation models (FMs)~\cite{jin2024debiased}, and large multimodal models (LMMs)~\cite{wu2024fmbench,jin2025see}. In parallel, medical AI systems are rapidly shifting toward \emph{agentic} architectures that solve tasks via \ul{\emph{multi-step pipelines}}, orchestrating multiple tools and models under the control of a large language model (LLM) planner~\cite{fallahpour2025medrax,wang2025survey,wang2025large}. By {dynamically selecting tools} such as classifiers, segmentators, report generators, and visual question answering (VQA) modules, agents can improve flexibility and interpretability, but they also introduce new pathways through which unfairness may arise and propagate.

In fact, \ul{\textit{fairness in agentic systems is not a direct extension of model-level fairness}}. Standalone fairness audits treat a model as a single decision function and measure disparities in final predictions. 
In contrast, agentic systems spend more effort on how to reason user queries, planning actions, and communicating among different models, resulting in \textit{new fairness failure modes} that do not exist in single-step models. 


However, current fairness evaluations rarely audit the agent’s internal process.
Without {process-level attribution}, it is hard to diagnose whether disparities are \emph{inherited} from particular vision-language tools, \emph{introduced} by the planner’s tool-selection and transition behavior, or \emph{amplified} during final response generation. 
We address this gap by conducting a \textit{stage-wise fairness decomposition} of tool-using chest X-ray agents for multiple-choice question answering. 

We study MedRAX~\cite{fallahpour2025medrax}, because it offers (i) \textit{complete trajectory observability} for auditing and decomposing, and (ii) a \textit{controlled single-agent pipeline} where decomposition into \textit{tool-exposure}, \textit{tool-transition}, and \textit{reasoning} disparities is well-defined, and intervention targets are modular.
Using CheXAgentBench and our curated MIMIC-FairnessVQA dataset, we evaluate fairness both at the \textit{outcome level} and at the \textit{process level}, decomposing agent unfairness into 
(i) \textit{tool-exposure bias}, 
(ii) \textit{tool-transition bias}, 
and (iii) \textit{LLM reasoning bias}. 
This decomposition yields a straightforward map from observed disparity to the agent stage most responsible, directly motivating our evaluation framework in Sec.~\ref{sec:method}.

\ul{\emph{Our contributions are threefold.}}
(1) We perform the first systematic demographic fairness evaluation of MedRAX-style chest X-ray agents across five driver LLMs under a unified setup.
(2) We propose \ours\ (\textbf{D}ecomposing \textbf{U}nfairness in \textbf{C}hest \textbf{X}-ray agents), a stage-wise framework with metrics that attribute disparities to \emph{tool exposure}, \emph{tool transition}, and \emph{LLM reasoning}.
(3) We curate \textit{MIMIC-FairnessVQA}, a demographic-aware benchmark with standardized (image, multi-choice question, demographics) instances for chest X-ray agents.

\section{\ours}

\begin{figure}[htbp]
    \centering
    \includegraphics[width=\linewidth]{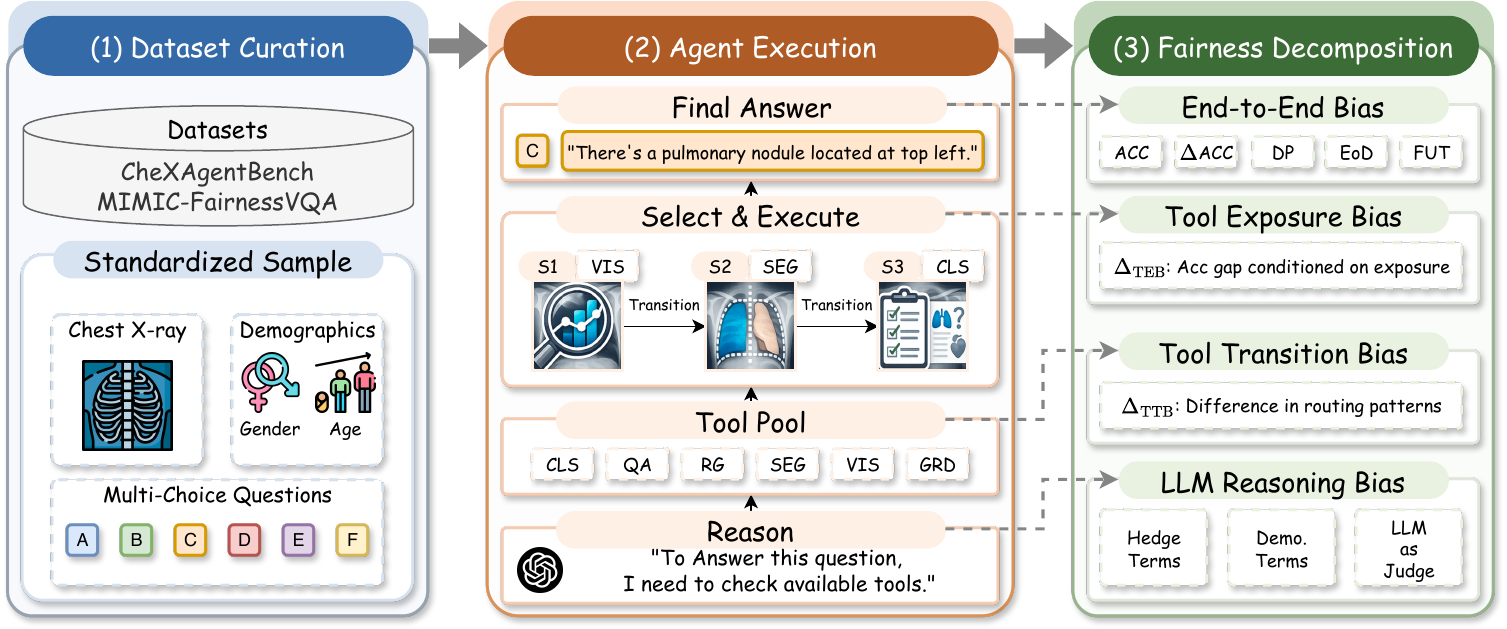}
    \caption{\textbf{Overview of \ours.} (1) Dataset curation: we use CheXAgentBench and curate MIMIC-FairnessVQA into a standardized form. (2) MedRAX agent execution (ReAct): a driver LLM iteratively reasons, selects tools, and synthesizes a final answer, where we highlight multiple points where bias can be introduced (tool exposure, tool transitions, and final synthesis). (3) Fairness decomposition: We evaluate \emph{end-to-end fairness} (ACC, $\Delta$ACC, DP, EoD, FUT) and decompose it into \emph{tool exposure bias} (gaps conditioned on tool presence), \emph{tool transition bias} (gaps in tool routing), and \emph{LLM reasoning bias} (gaps in synthesis behaviors).}
    \label{fig:teaser}
\end{figure}


\subsection{Preliminaries}
\noindent\textbf{Agentic systems in medical imaging.} Traditional \emph{standalone} medical imaging models map an input image directly to an \emph{end-to-end output} (e.g., a label or report), so fairness evaluation typically concerns only the final prediction~\cite{xu2024addressing,jin2024fairmedfm}. 
In contrast, recent systems are increasingly \emph{agentic}~\cite{fallahpour2025medrax,sharma2024cxr,kim2024mdagents}: an LLM planner decomposes a query into multiple steps and autonomously invokes specialized tools (e.g., classifiers, segmenters, report generators, grounding, visualization, or VQA modules) to produce the final answer, without human intervention as shown in Fig.~\ref{fig:teaser}. 
For example, MedRAX~\cite{fallahpour2025medrax} is a representative tool-using X-ray agent framework that follows a ReAct-style loop~\cite{yao2022react}: the agent iteratively \emph{reasons} about the next step, \emph{acts} by calling a tool or generating an intermediate response, and \emph{observes} the result to update its context until completion. 
This structured execution trace \emph{exposes multiple decision points beyond the final output}, enabling the motivation for stage-wise analysis and decomposition of unfairness in agentic systems.


\noindent\textbf{Fairness in medical imaging.} Prior work on fairness in medical imaging has largely evaluated \emph{standalone} predictors, operationalizing equity as comparable performance and error behavior across demographic groups~\cite{zhang2022improving,jackson2025enhancement,seyyed2021underdiagnosis,larrazabal2020gender,glocker2023risk}. \emph{Group fairness} measures each subgroup's \textit{end-to-end performance} (e.g., accuracy) and summarizes disparity by the \emph{gap} between groups~\cite{hardt2016equality};
These notions are well-suited to single-step models, but emerging \emph{agentic} pipelines produce answers via intermediate decisions (e.g., tool exposure, tool transition, and LLM synthesis), creating additional pathways for disparities that are not visible from final outcomes alone. This motivates auditing fairness not only at the outcome level, but also at the \emph{process level}: whether different groups induce systematically different tool-routing policies or generation behaviors that may ultimately drive end-to-end gaps.

\subsection{Problem Setting and Agent Execution}


Let $\mathcal{D}=\{(x_i, q_i, y_i, a_i)\}_{i=1}^N$ denote a dataset of chest X-ray questions, where $x$ is an image (and optional clinical context), $q$ is a question with $K$ candidate answers, $y \in \{1,\ldots,K\}$ is the ground-truth option index, and $a$ is a sensitive attribute (e.g., gender). Let $g \in \mathcal{G}$ index sensitive subgroups induced by $a$ (e.g., $\mathcal{G}=\{\text{Female},\text{Male}\}$ or $\{\text{Age}<60,\text{Age}\ge60\}$).
An agent produces a final prediction $\hat{y}$ by executing a tool-augmented trajectory $\tau$:
\begin{equation}
\tau = \big( s_0, (t_1,o_1), \ldots, (t_T,o_T), \hat{y} \big),
\end{equation}
where $s_0$ is the initial state formed from $(x,q)$, $t_j \in \mathcal{T}$ denotes the selected tool at step $j$ from a tool set $\mathcal{T}$, $o_j$ is the corresponding tool output (observation), and $T$ is the number of tool calls before producing the final answer. We assume access to execution logs containing $(t_j,o_j)$ and the final natural-language response.

\noindent\textbf{Driver LLMs.} Five LLMs including LLaMA3.1-8B~\cite{grattafiori2024llama}, Ministral-3-8B~\cite{liu2026ministral}, Qwen3VL-8B~\cite{bai2025qwen3}, Qwen3-8B~\cite{yang2025qwen3}, and Gemini3-Flash~\cite{comanici2025gemini}, are used to drive the agent, to evaluate potential unfairness introduced by the driver LLM.

\noindent\textbf{Tools Pool.} Six categories of tools are included in the tools pool for calling, including classifier (CLS), visual question answering module (QA), report generator (RG), segmentator (SEG), visualizer (VIS), and phrase grounding module (GRD).




\subsection{Fairness Decomposition}
\label{sec:method}

\noindent\textbf{End-to-end bias.}
Three types of metrics are adopted following~\cite{jin2024fairmedfm}, i.e., (i) \ul{\emph{Utility measurement}}: accuracy ($\text{ACC}$); (ii) \ul{\emph{Fairness measurement}}: delta accuracy ($\Delta\text{ACC}$), demographic parity (DP), Equalized Odds (EoD)~\cite{hardt2016equality}; (iii) \ul{\emph{Trade-off measurement}}: fairness-utility trade-off score (FUT)~\cite{jin2024fairmedfm}.

\noindent\textbf{Tool-exposure bias.} \ul{\textit{Motivation:}} even when the agent uses the \emph{same} tool, that tool may deliver
\textit{unequal downstream utility across groups} (e.g., due to training imbalance),
which then propagates to the final decision. \ul{\textit{Definition:}} for a tool $A$, let $E_A(\tau)=\mathbbm{1}[A\in\tau]$ indicate whether $A$ is used at least once.
We define tool-exposure bias as the \textit{subgroup accuracy gap conditioned on exposure}:
\begin{equation}
\Delta_{\mathrm{TEB}}(A)
=
\mathrm{Acc}\!\left(g{=}g_1 \mid E_A(\tau){=}1\right)
-
\mathrm{Acc}\!\left(g{=}g_2 \mid E_A(\tau){=}1\right).
\end{equation}
\ul{\textit{Evaluation:}} we filter to instances with $E_A(\tau)=1$, compute subgroup accuracies,
and report $\Delta_{\mathrm{TEB}}(A)$ (and $|\Delta_{\mathrm{TEB}}(A)|$).
\textit{To avoid confounding with routing}, we also report each group's \textit{exposure rate}
$P(E_A(\tau)=1\mid g)$.




\noindent\textbf{Tool-transition bias.} \ul{\textit{Motivation:}} unfairness can arise even with identical tools if the LLM planner
\textit{routes different demographics through systematically different tool chains}
(e.g., longer or less reliable sequences example in the introduction). \ul{\textit{Definition:}} let the tools pool be $\mathcal{T}=\{A_1,\ldots,A_T\}$.
For each subgroup $g$, we estimate a \textit{Markov transition matrix} $P^{(g)}\in\mathbb{R}^{T\times T}$:
\begin{equation}
p^{(g)}_{i,j} = P(t_{k+1}=A_j \mid t_k=A_i, g),
\qquad \sum_{j=1}^{T} p^{(g)}_{i,j}=1.
\end{equation}
We define tool-transition bias as the \textit{difference in routing patterns}:
\begin{equation}
\Delta_{\mathrm{TTB}} = P^{(g_1)} - P^{(g_2)}.
\end{equation}
\ul{\textit{Evaluation:}} entries of $\Delta_{\mathrm{TTB}}$ identify which tool-to-tool transitions are
\textit{more frequent for one group}, highlighting planning-level disparity beyond final accuracy.


\noindent\textbf{LLM reasoning bias.} \ul{\textit{Motivation:}} even with identical trajectories, unfairness may manifest even when tools and answers are comparable, because the driver LLM can produce \textit{demographic-dependent reasoning quality and communication style} (e.g., different uncertainty expression or demographic framing). We therefore decompose \textit{synthesis-level disparities} directly from the agent's final response text. \ul{\textit{Definition:}} For a sensitive attribute with groups $\mathcal{G}$ and a response-level feature $f$, we compute subgroup means $\mu_g=\mathbb{E}[f\mid g]$ and define the reasoning-bias gap as:
\begin{equation}
\Delta f=\max_{g\in\mathcal{G}} \mu_g - \min_{g\in\mathcal{G}} \mu_g .
\end{equation}
\ul{\textit{Evaluation:}} We report three complementary gaps:
(i) \textit{JudgeGap}: $\Delta\text{JudgeGap}$, where $f$ is a \textit{reasoning quality score} produced by an external LLM judge given the \textit{question} and the agent’s final response. We adapt the exact prompt in ~\cite{zheng2023judging} for the LLM judge.
(ii) \textit{Hedge}~\cite{agarwal2010detecting,kim2024m}: $\Delta\text{Hedge}$, where $f$ counts \textit{hedging cues} (e.g., \emph{may, might, possibly, likely, appears}) via case-insensitive pattern matching;  
(iii) \textit{Demographic}~\cite{derner2025leveraging,hanna2025assessing}: $\Delta\text{Demo.}$, where $f$ counts \textit{explicit demographic terms} (e.g., \emph{male, female, elderly, young}) via pattern matching.
Above, JudgeGap captures overall reasoning quality disparity, while $\Delta$Hedge and $\Delta$Demo. capture systematic differences in uncertainty expression and demographic framing.

\subsection{Dataset Curation}

Evaluating fairness in \emph{agentic} chest X-ray systems requires \emph{agent-suitable} questions that support multi-step tool use, include sensitive attributes, and allow automatic scoring. Such datasets are scarce; to our knowledge, only ChestAgentBench~\cite{fallahpour2025medrax} provides both agent-oriented questions and patient demographics. We therefore use ChestAgentBench and curate a new MIMIC-based benchmark.

\noindent\textbf{CheXAgentBench~\cite{fallahpour2025medrax}.} It contains 2,500 diagnosis questions from 675 expert-curated clinical cases (Eurorad\footnote{\url{https://www.eurorad.org/}}). Each sample includes a chest X-ray, patient demographics, and a six-choice question. The dataset is available via HuggingFace~\cite{fallahpour_chestagentbench}.

\noindent\textbf{MIMIC-FairnessVQA\footnote{Dataset is provided \href{https://anonymous.4open.science/r/DUCK-E5FE/README.md}{here}.}.}
We curate MIMIC-FairnessVQA from MIMIC-CXR~\cite{johnson2019mimic} to enable demographic-aware evaluation at scale:
(i) \ul{\emph{Demographics-balanced sampling}}: we sample 400 studies balanced by gender and age (threshold 60);
(ii) \ul{\emph{Information extraction}}: we extract metadata, the final report, and the X-ray image;
(iii) \ul{\emph{LLM-based question generation}}: we prompt LLM (using a MedRAX-style template) to generate six-choice questions with explanations across five task focuses per study, yielding 2,000 instances.
We manually verify answer correctness.

\section{Results}

In this section, we systematically inspect unfairness in agent instruction following the scheme in Sec.~\ref{sec:method}, including end-to-end bias, tool-exposure bias, tool-transition bias, and LLM-reasoning bias.

\noindent\textbf{Implementation details.} We follow the open-source MedRAX codebase for evaluation and use DeepSeek as the judge model. We assess statistical significance with non-parametric bootstrapping (1,000 resamples) and report confidence intervals (CIs). For sensitive attributes, we consider gender (male vs.\ female) and age (young $<$60 vs.\ old $\ge$60).




\begin{table}[htbp]
    \centering
    \caption{\textbf{Overview of the end-to-end fairness performance.} $\text{Mean}_{\text{CI}}$ is reported. All values are in percentage. \xbest{Best} and \xsecond{Second Best} in each group are highlighted.}
    \label{tab:fairness_metrics}
    \resizebox{\linewidth}{!}{
    \begin{tabular}{ll|rrrrr|rrrrr}
    \toprule
    \multicolumn{2}{c|}{Configuration} &\multicolumn{5}{c|}{CheXAgentBench} & \multicolumn{5}{c}{MIMIC-FairnessVQA} \\
     Attr. & LLM & ACC $\uparrow$ & $\Delta$ACC $\downarrow$& DP $\downarrow$& EoD$\downarrow$ & FUT $\uparrow$& ACC $\uparrow$ & $\Delta$ACC $\downarrow$& DP $\downarrow$& EoD$\downarrow$ & FUT $\uparrow$\\
        \cline{1-2} \cline{3-7}\cline{8-12}
    \multirow{5}{*}{Gender} & 
  \textbf{LLaMA3.1} & $50.83_{[48.88, 52.76]}$ & $\xbest{1.53_{[0.06, 4.45]}}$ & $5.04_{[2.32, 7.93]}$ & $15.10_{[5.63, 26.98]}$ & $50.29_{[48.20, 52.30]}$ & $32.89_{[30.85, 34.95]}$ & $2.00_{[0.10, 5.86]}$ & $\xbest{2.99_{[1.12, 6.28]}}$ & $11.45_{[5.83, 18.97]}$ & $32.43_{[30.39, 34.56]}$ \\
    & \textbf{Ministral-3} & $51.48_{[49.72, 53.36]}$ & $\xsecond{1.68_{[0.06, 4.40]}}$ & $\xbest{3.66_{[1.48, 6.73]}}$ & $10.48_{[4.12, 19.60]}$ & $50.85_{[48.75, 52.88]}$& $29.42_{[27.35, 31.50]}$ & $2.30_{[0.11, 6.07]}$ & $3.35_{[1.39, 5.86]}$ & $\xsecond{11.21_{[5.39, 18.36]}}$ & $28.95_{[26.86, 31.07]}$ \\
    &\textbf{Qwen3VL} & $\xsecond{67.26_{[65.11, 69.28]}}$ & $3.21_{[0.21, 7.11]}$ & $4.47_{[1.42, 8.12]}$ & $11.28_{[4.87, 22.64]}$ & $\xsecond{65.98_{[63.41, 68.45]}}$& $\xbest{54.75_{[52.07, 57.39]}}$ & $2.17_{[0.11, 6.29]}$ & $4.22_{[1.74, 7.56]}$ & $11.47_{[4.93, 20.61]}$ & $\xbest{53.92_{[51.02, 56.77]}}$ \\
    &\textbf{Qwen3} & $\xbest{69.10_{[67.24, 70.92]}}$ & $1.92_{[0.09, 5.07]}$ & $\xsecond{3.72_{[1.52, 6.62]}}$ & $\xbest{9.16_{[3.45, 19.62]}}$ & $\xbest{68.09_{[65.74, 70.32]}}$& $49.89_{[47.55, 52.10]}$ & $\xbest{1.90_{[0.09, 5.30]}}$ & $\xsecond{3.11_{[1.14, 5.98]}}$ & $11.32_{[5.06, 19.56]}$ & $\xsecond{49.23_{[46.77, 51.37]}}$ \\
    &\textbf{Gemini3} & $50.60_{[48.64, 52.64]}$ & $3.62_{[0.36, 7.47]}$ & $4.99_{[2.61, 7.99]}$ & $\xsecond{9.56_{[4.83, 16.61]}}$ & $49.15_{[46.82, 51.59]}$& $49.69_{[47.50, 51.95]}$ & $\xsecond{1.99_{[0.08, 5.31]}}$ & $3.13_{[1.36, 5.38]}$ & $\xbest{11.01_{[5.31, 19.10]}}$ & $49.01_{[46.54, 51.48]}$ \\
        \cline{1-2} \cline{3-7}\cline{8-12}
     \multirow{5}{*}{Age} &
    \textbf{LLaMA3.1} & $50.83_{[48.88, 52.76]}$ & $2.66_{[0.13, 6.45]}$ & $\xbest{2.92_{[1.00, 5.69]}}$ & $13.58_{[5.91, 23.11]}$ & $49.47_{[46.60, 51.87]}$& $32.89_{[30.85, 34.95]}$ & $\xsecond{2.27_{[0.08, 6.00]}}$ & $\xbest{2.82_{[1.22, 5.76]}}$ & $\xbest{9.63_{[4.48, 16.65]}}$ & $32.40_{[30.30, 34.53]}$ \\ 
        & \textbf{Ministral-3} & ${51.48_{[49.72, 53.36]}}$ & $\xbest{1.89_{[0.07, 5.33]}}$ & $\xsecond{3.00_{[1.16, 5.69]}}$ & ${12.08_{[4.85, 21.98]}}$ & ${50.60_{[48.02, 52.81]}}$& $29.42_{[27.35, 31.50]}$ & $3.55_{[0.22, 7.50]}$ & $4.73_{[2.32, 7.70]}$ & $\xsecond{11.38_{[5.62, 18.52]}}$ & $28.65_{[26.53, 30.74]}$ \\
    &\textbf{Qwen3VL} & $\xsecond{67.26_{[65.11, 69.28]}}$ & $\xsecond{2.10_{[0.06, 5.99]}}$ & $4.27_{[1.74, 7.52]}$ & $20.79_{[8.91, 33.72]}$ & $\xsecond{66.14_{[62.90, 68.65]}}$& $\xbest{54.75_{[52.07, 57.39]}}$ & $\xbest{2.18_{[0.11, 6.19]}}$ & $5.24_{[2.55, 9.03]}$ & $16.89_{[7.96, 26.50]}$ & $\xbest{53.94_{[50.95, 56.81]}}$ \\
    &\textbf{Qwen3} & $\xbest{69.10_{[67.24, 70.92]}}$ & $2.56_{[0.14, 6.12]}$ & $3.19_{[1.37, 5.95]}$ & $\xbest{10.98_{[5.13, 19.79]}}$ & $\xbest{67.45_{[64.64, 69.99]}}$ & $\xsecond{49.89_{[47.55, 52.10]}}$ & $2.34_{[0.14, 6.29]}$ & $5.15_{[2.75, 8.23]}$ & ${13.00_{[6.26, 21.63]}}$ & $\xsecond{49.11_{[46.48, 51.44]}}$ \\
    &\textbf{Gemini3} & $50.60_{[48.64, 52.64]}$ & $2.37_{[0.14, 6.01]}$ & $3.15_{[1.38, 5.28]}$ & $\xsecond{9.89_{[4.89, 17.24]}}$ & $49.41_{[46.59, 51.94]}$& $49.69_{[47.50, 51.95]}$ & $2.54_{[0.14, 6.60]}$ & $\xsecond{4.14_{[2.05, 6.49]}}$ & $12.96_{[6.46, 20.47]}$ & $48.85_{[46.35, 51.13]}$  \\
    \bottomrule
    \end{tabular}
    }
    
\end{table}

\noindent\textbf{End-to-end bias.}
We firstly compare the end-to-end fairness of different LLMs across gender and age attributes in Table~\ref{tab:fairness_metrics}. \emph{All LLMs exhibit varying degrees of unfairness across the two datasets.}\footnote{Directly applying Qwen3VL-8B and Gemini3-Flash as end-to-end tools, rather than agents, results in poor overall utility.}
In terms of ACC, the Qwen3 model performs exceptionally well on both datasets, achieving 69.10\% and 49.89\% respectively, while maintaining relatively low $\Delta$ACC and DP values, indicating strong fairness. The Qwen3VL also demonstrates high accuracy, but its higher EoD in the Age group suggests some fairness risks in that dimension. Overall, the Qwen series achieves a better balance between fairness and utility (FUT).
Note that DP is closely tied to the base rates of individual subgroups; therefore, careful examination is required before adopting this metric.

\begin{figure}[htbp]
    \centering
    \includegraphics[width=0.9\linewidth]{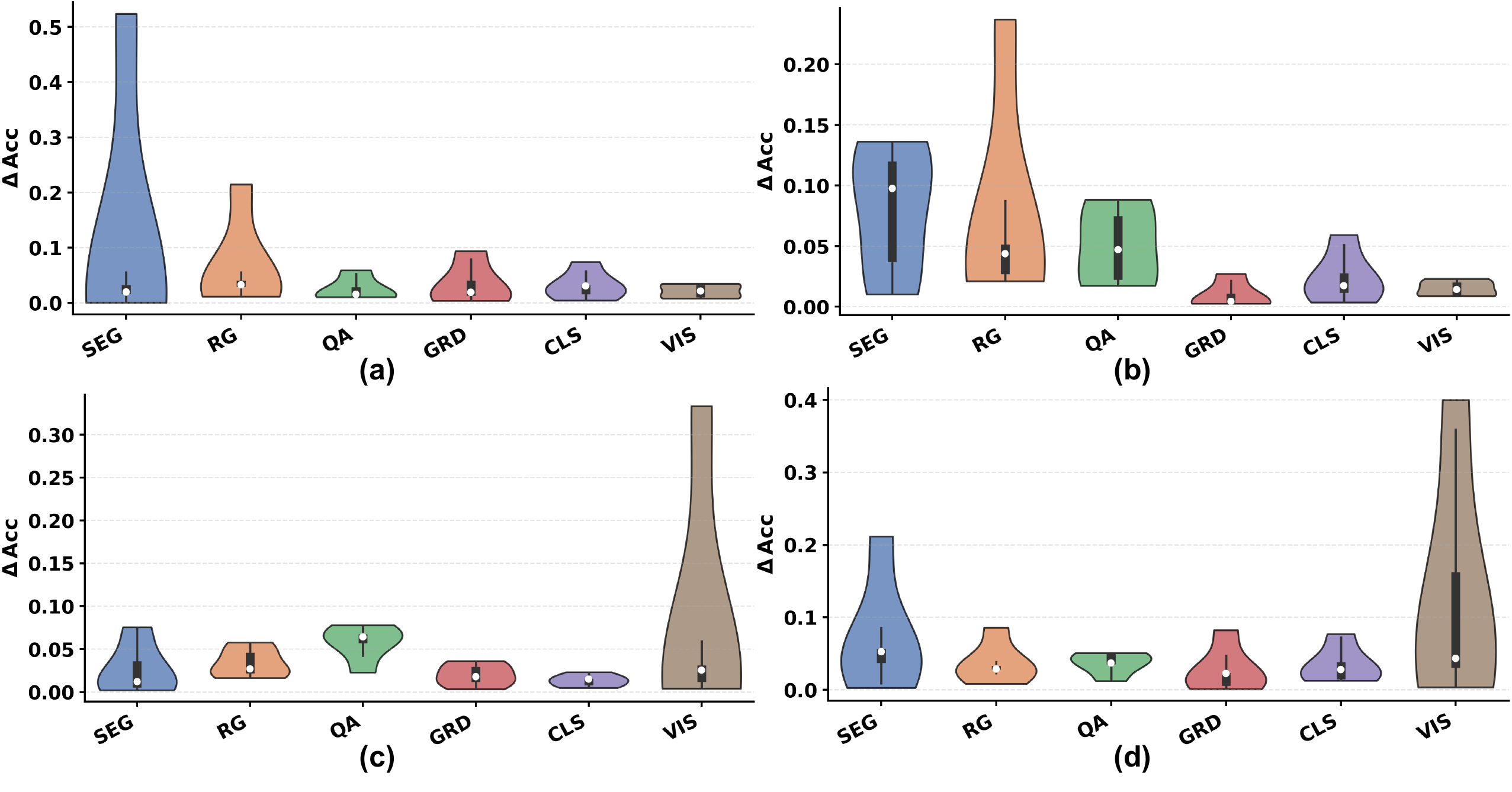}
    \caption{\textbf{Tool exposure bias across tools and subgroups.} (a) Gender on CheXAgentBench, (b) Age on CheXAgentBench, (c) Gender on MIMIC-FairnessVQA, and (d) Age on MIMIC-FairnessVQA.}
    \label{fig:tool-bias}
\end{figure}

\noindent\textbf{Tool-exposure bias.} Fig.~\ref{fig:tool-bias} summarizes \emph{tool exposure} accuracy gaps: for each tool, the violin shows the distribution (across driver LLMs) of $|\Delta\mathrm{ACC}|$ computed on instances where the tool is invoked. Two patterns stand out. 
\emph{(i) Tool dependence:} on CheXAgentBench, \emph{segmentation} exhibits the largest and heaviest-tailed gaps (especially for gender), with \emph{report generation} a distant second. In contrast, \emph{classifier} and \emph{grounding} remain consistently near zero. 
\emph{(ii) Dataset and attribute shift:} on MIMIC-FairnessVQA, the largest disparities concentrate in the \emph{visualizer}, suggesting that different tool outputs become the main ``fairness bottleneck'' under different data distributions. Connecting to the end-to-end results (Table~\ref{tab:fairness_metrics}), the overall $\Delta\mathrm{ACC}$ gaps are smaller than the worst exposure-conditioned gaps because they average over heterogeneous trajectories and are further shaped by \emph{transition bias} and \emph{LLM reasoning bias}. Nevertheless, the exposure analysis isolates which tools are most likely to seed end-to-end unfairness, providing a mechanistic explanation for why some backbones exhibit larger $\Delta\mathrm{ACC}$ even when overall accuracy is comparable.

\begin{figure}[htbp]
    \centering
    \includegraphics[width=\linewidth]{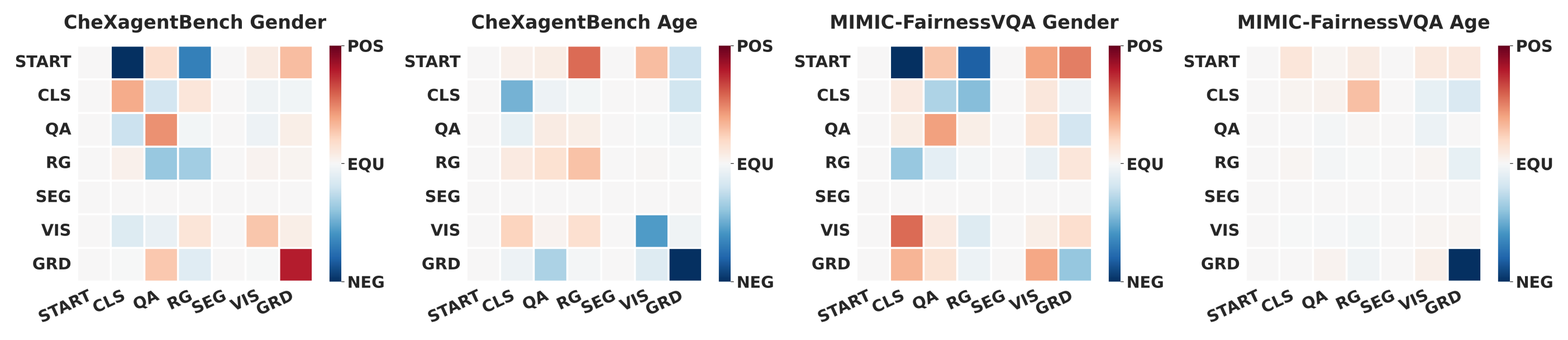}
    \caption{\textbf{Tool transition biases represented as matrices.} START: start indicator; CLS: classifier; RG: report generator; SEG: segmentator; VIS: visualizer; GRD: phrase grounding. Cells in red denote values larger than zero, while cells in blue denote values smaller than zero.$\Delta_{\text{TTB}}^{\text{gender}} = P^{\text{male}} - P^{\text{female}}$, $\Delta_{\text{TTB}}^{\text{age}} = P^{\text{young}} - P^{\text{old}}$. All subfigures present results averaged across different LLMs.}
    \label{fig:transition-mat}
\end{figure}

\noindent\textbf{Tool-transition bias.}
As shown in Fig.~\ref{fig:transition-mat}, distinct tool transition patterns are observed across different attributes in various datasets, with gender-based differences being more pronounced than those related to age. In both datasets, female patients are more likely than male patients to proceed directly from the \emph{Classifier} or \emph{Report Generator}.
Within the age attribute of CheXAgentBench, younger individuals exhibit a similar tendency compared to older ones. 
An interesting observation in MIMIC-FairnessVQA is that male patients often call the \emph{Classifier} again after using the \emph{Visualizer}, which is a pattern that warrants further investigation. 
Additionally, in both the age attribute of CheXAgentBench and MIMIC-FairnessVQA, older individuals and male patients show a higher frequency of repeated calls to \emph{Grounding} tools, which may suggest more effort is needed for answering questions about these demographics.

\begin{table}[htbp]
    \centering
    \caption{\textbf{Subgroup gaps of reasoning-bias.} $\text{Mean}_{\text{CI}}$ is reported. All values are multiplied by 100 for better readability. \xbest{Best} and \xsecond{Second Best} are highlighted.}
    \label{tab:reasoning-bias}
    \resizebox{\linewidth}{!}{
    \begin{tabular}{ll|rrr|rrr}
    \toprule
    \multicolumn{2}{c|}{Configuration} & \multicolumn{3}{c|}{CheXAgentBench} & \multicolumn{3}{c}{MIMIC-FairnessVQA} \\
    Attr. & LLM & $\Delta$JudgeGap$\downarrow$ & $\Delta$Hedge$\downarrow$ & $\Delta$Demo.$\downarrow$ & $\Delta$JudgeGap$\downarrow$ & $\Delta$Hedge$\downarrow$ & $\Delta$Demo.$\downarrow$\\
        \cline{1-2} \cline{3-5}\cline{6-8}
    \multirow{5}{*}{Gender} & \textbf{LLaMA3.1} 
    & $6.19_{[0.21,19.00]}$ 
    & \xsecond{$4.97_{[0.25,11.66]}$} 
    & \xsecond{$0.09_{[0.02,1.66]}$} 
    & \xbest{$0.10_{[0.30,21.70]}$} 
    & \xbest{$3.10_{[0.20,12.40]}$} 
    & \xsecond{$0.10_{[0.00,0.30]}$} \\
    
    & \textbf{Ministral-3} 
    & \xsecond{$2.86_{[0.15,8.92]}$} 
    & $12.20_{[0.63,31.73]}$ 
    & $0.56_{[0.06,3.74]}$ 
    & $3.10_{[0.40,23.40]}$ 
    & $4.90_{[0.60,44.40]}$ 
    & $0.20_{[0.00,1.20]}$ \\
    
    & \textbf{Qwen3VL} 
    & $4.59_{[0.22,11.39]}$ 
    & $534.95_{[3.58,1438.18]}$ 
    & $1.57_{[0.08,4.21]}$ 
    & $21.67_{[2.66,41.90]}$ 
    & $831.10_{[44.75,3489.64]}$ 
    & $0.10_{[0.00,0.30]}$ \\
    
    & \textbf{Qwen3} 
    & $5.58_{[0.47,11.69]}$ 
    & $12.58_{[0.65,51.20]}$ 
    & $3.63_{[0.57,6.88]}$ 
    & \xsecond{$1.90_{[0.30,24.80]}$} 
    & \xsecond{$4.40_{[1.00,69.20]}$} 
    & \xbest{$0.00_{[0.00,0.00]}$} \\
    
    & \textbf{Gemini3} 
    & \xbest{$1.67_{[0.09,4.48]}$} 
    & \xbest{$0.89_{[0.13,11.09]}$} 
    & \xbest{$0.01_{[0.06,4.22]}$} 
    & $21.20_{[10.49,32.50]}$ 
    & $10.40_{[0.80,25.00]}$ 
    & $0.30_{[0.00,1.30]}$ \\
    
    \cline{1-2} \cline{3-5}\cline{6-8}
    \multirow{5}{*}{Age} & \textbf{LLaMA3.1} 
    & $2.60_{[0.29,16.73]}$ 
    & \xsecond{$5.94_{[0.51,12.88]}$} 
    & \xbest{$1.24_{[0.08,2.71]}$} 
    & $22.90_{[4.99,41.11]}$ 
    & \xbest{$1.30_{[0.20,11.10]}$} 
    & \xsecond{$0.10_{[0.00,0.30]}$} \\
    
    & \textbf{Ministral-3} 
    & $5.23_{[0.45,11.47]}$ 
    & $26.29_{[5.16,50.15]}$ 
    & $4.06_{[1.61,6.54]}$ 
    & $28.10_{[8.70,47.70]}$ 
    & $56.70_{[18.50,94.10]}$ 
    & $0.80_{[0.10,1.80]}$ \\
    
    & \textbf{Qwen3VL} 
    & $6.65_{[0.46,13.50]}$ 
    & $407.66_{[14.56,1080.75]}$ 
    & $3.12_{[1.01,5.53]}$ 
    & \xsecond{$15.74_{[1.16,35.15]}$} 
    & $1165.50_{[89.77,3893.21]}$ 
    & $0.10_{[0.00,0.30]}$ \\
    
    & \textbf{Qwen3} 
    & \xsecond{$1.10_{[0.10,7.75]}$} 
    & \xbest{$4.70_{[0.56,47.52]}$} 
    & \xsecond{$2.14_{[0.11,4.94]}$} 
    & $24.10_{[4.70,44.60]}$ 
    & $67.20_{[9.69,127.11]}$ 
    & \xbest{$0.00_{[0.00,0.00]}$} \\
    
    & \textbf{Gemini3} 
    & \xbest{$0.14_{[0.05,3.16]}$} 
    & $5.88_{[0.27,17.05]}$ 
    & $8.26_{[5.08,11.48]}$ 
    & \xbest{$2.40_{[0.30,13.80]}$} 
    & \xsecond{$17.40_{[5.00,31.01]}$} 
    & $0.50_{[0.00,1.50]}$ \\
    \bottomrule
    \end{tabular}
    }
\end{table}

\noindent\textbf{LLM-reasoning bias.} 
Table~\ref{tab:reasoning-bias} reports subgroup gaps in three response-level features: judge-scored reasoning quality ($\Delta$JudgeGap), hedging frequency ($\Delta$Hedge), and explicit demographic mentions ($\Delta$Demo). Two trends are clear. 
\emph{First,} reasoning bias gaps are highly model-dependent: Qwen3VL shows larger hedging gaps than all other backbones across both datasets and attributes, indicating substantially different uncertainty expression between subgroups. In contrast, LLaMA3.1 and Qwen3 generally exhibit low-to-moderate hedging gaps, while Gemini3 is consistently among the lowest on CheXAgentBench gender. 
\emph{Second,} the three measures capture distinct behaviors and need not move together: some models achieve low demographic-term gaps ($\Delta$Demo) yet still show large hedging or judge gaps (e.g., Qwen3VL on MIMIC-FairnessVQA), suggesting that subgroup differences primarily manifest in uncertainty style rather than explicit demographic framing. Overall, these results highlight that driver LLMs can introduce substantial subgroup-dependent variation in response synthesis even when the same questions and tool outputs are provided.
\section{Conclusion}
We audit demographic fairness in MedRAX-style tool-using chest X-ray agents and propose \ours\ to decompose end-to-end disparities into \emph{tool exposure}, \emph{tool transition}, and \emph{LLM reasoning} biases. Across five driver LLMs and two agent-suitable benchmarks (CheXAgentBench and our MIMIC-FairnessVQA), we find persistent subgroup gaps and show that unfairness can originate from intermediate tool use, routing, and response synthesis, not only final predictions. Our results motivate process-level fairness evaluation for agentic medical imaging systems, and future work will develop stage-targeted mitigation and extend audits to broader clinical tasks and agent designs.

\begin{credits}

\subsubsection{\discintname}
The authors have no competing interests to declare that are relevant to the content of this article. 
\end{credits}

\bibliographystyle{splncs04}
\bibliography{reference}

@article{jin2024fairmedfm,
  title={Fairmedfm: fairness benchmarking for medical imaging foundation models},
  author={Jin, Ruinan and Xu, Zikang and Zhong, Yuan and Yao, Qingsong and QI, DOU and Zhou, S Kevin and Li, Xiaoxiao},
  journal={Advances in Neural Information Processing Systems},
  volume={37},
  pages={111318--111357},
  year={2024}
}

@misc{fallahpour_chestagentbench,
  author       = {Fallahpour, Alireza and others},
  title        = {ChestAgentBench},
  year         = {2025}, 
  howpublished = {Hugging Face Dataset},
  url          = {https://huggingface.co/datasets/wanglab/chest-agent-bench},
}

@article{liu2023translational,
  title={A translational perspective towards clinical AI fairness},
  author={Liu, Mingxuan and Ning, Yilin and Teixayavong, Salinelat and Mertens, Mayli and Xu, Jie and Ting, Daniel Shu Wei and Cheng, Lionel Tim-Ee and Ong, Jasmine Chiat Ling and Teo, Zhen Ling and Tan, Ting Fang and others},
  journal={NPJ digital medicine},
  volume={6},
  number={1},
  pages={172},
  year={2023},
  publisher={Nature Publishing Group UK London}
}

@article{zhou2021review,
  title={A review of deep learning in medical imaging: Imaging traits, technology trends, case studies with progress highlights, and future promises},
  author={Zhou, S Kevin and Greenspan, Hayit and Davatzikos, Christos and Duncan, James S and Van Ginneken, Bram and Madabhushi, Anant and Prince, Jerry L and Rueckert, Daniel and Summers, Ronald M},
  journal={Proceedings of the IEEE},
  volume={109},
  number={5},
  pages={820--838},
  year={2021},
  publisher={IEEE}
}

@article{wu2024fmbench,
  title={Fmbench: Benchmarking fairness in multimodal large language models on medical tasks},
  author={Wu, Peiran and Liu, Che and Chen, Canyu and Li, Jun and Bercea, Cosmin I and Arcucci, Rossella},
  journal={arXiv preprint arXiv:2410.01089},
  year={2024}
}

@inproceedings{
fallahpour2025medrax,
title={Med{RAX}: Medical Reasoning Agent for Chest X-ray},
author={Adibvafa Fallahpour and Jun Ma and Alif Munim and Hongwei Lyu and BO WANG},
booktitle={Forty-second International Conference on Machine Learning},
year={2025},
url={https://openreview.net/forum?id=JiFfij5iv0}
}

@inproceedings{wang2025survey,
    title = "A Survey of {LLM}-based Agents in Medicine: How far are we from Baymax?",
    author = "Wang, Wenxuan  and
      Ma, Zizhan  and
      Wang, Zheng  and
      Wu, Chenghan  and
      Ji, Jiaming  and
      Chen, Wenting  and
      Li, Xiang  and
      Yuan, Yixuan",
    editor = "Che, Wanxiang  and
      Nabende, Joyce  and
      Shutova, Ekaterina  and
      Pilehvar, Mohammad Taher",
    booktitle = "Findings of the Association for Computational Linguistics: ACL 2025",
    month = jul,
    year = "2025",
    address = "Vienna, Austria",
    publisher = "Association for Computational Linguistics",
    url = "https://aclanthology.org/2025.findings-acl.539/",
    doi = "10.18653/v1/2025.findings-acl.539",
    pages = "10345--10359",
    ISBN = "979-8-89176-256-5"
}

@article{wang2025large,
  title={Large Language Model for Medical Images: A Survey of Taxonomy, Systematic Review, and Future Trends},
  author={Wang, Peng and Lu, Wenpeng and Lu, Chunlin and Zhou, Ruoxi and Li, Min and Qin, Libo},
  journal={Big Data Mining and Analytics},
  volume={8},
  number={2},
  pages={496--517},
  year={2025},
  publisher={TUP}
}

@article{agarwal2010detecting,
  title={Detecting hedge cues and their scope in biomedical text with conditional random fields},
  author={Agarwal, Shashank and Yu, Hong},
  journal={Journal of biomedical informatics},
  volume={43},
  number={6},
  pages={953--961},
  year={2010},
  publisher={Elsevier}
}

@article{grattafiori2024llama,
  title={The llama 3 herd of models},
  author={Grattafiori, Aaron and Dubey, Abhimanyu and Jauhri, Abhinav and Pandey, Abhinav and Kadian, Abhishek and Al-Dahle, Ahmad and Letman, Aiesha and Mathur, Akhil and Schelten, Alan and Vaughan, Alex and others},
  journal={arXiv preprint arXiv:2407.21783},
  year={2024}
}

@article{liu2026ministral,
  title={Ministral 3},
  author={Liu, Alexander H and Khandelwal, Kartik and Subramanian, Sandeep and Jouault, Victor and Rastogi, Abhinav and Sad{\'e}, Adrien and Jeffares, Alan and Jiang, Albert and Cahill, Alexandre and Gavaudan, Alexandre and others},
  journal={arXiv preprint arXiv:2601.08584},
  year={2026}
}

@article{bai2025qwen3,
  title={Qwen3-vl technical report},
  author={Bai, Shuai and Cai, Yuxuan and Chen, Ruizhe and Chen, Keqin and Chen, Xionghui and Cheng, Zesen and Deng, Lianghao and Ding, Wei and Gao, Chang and Ge, Chunjiang and others},
  journal={arXiv preprint arXiv:2511.21631},
  year={2025}
}

@article{yang2025qwen3,
  title={Qwen3 technical report},
  author={Yang, An and Li, Anfeng and Yang, Baosong and Zhang, Beichen and Hui, Binyuan and Zheng, Bo and Yu, Bowen and Gao, Chang and Huang, Chengen and Lv, Chenxu and others},
  journal={arXiv preprint arXiv:2505.09388},
  year={2025}
}

@article{comanici2025gemini,
  title={Gemini 2.5: Pushing the frontier with advanced reasoning, multimodality, long context, and next generation agentic capabilities},
  author={Comanici, Gheorghe and Bieber, Eric and Schaekermann, Mike and Pasupat, Ice and Sachdeva, Noveen and Dhillon, Inderjit and Blistein, Marcel and Ram, Ori and Zhang, Dan and Rosen, Evan and others},
  journal={arXiv preprint arXiv:2507.06261},
  year={2025}
}

@inproceedings{yao2022react,
  title={React: Synergizing reasoning and acting in language models},
  author={Yao, Shunyu and Zhao, Jeffrey and Yu, Dian and Du, Nan and Shafran, Izhak and Narasimhan, Karthik R and Cao, Yuan},
  booktitle={The eleventh international conference on learning representations},
  year={2022}
}

@article{xu2024addressing,
  title={Addressing fairness issues in deep learning-based medical image analysis: a systematic review},
  author={Xu, Zikang and Li, Jun and Yao, Qingsong and Li, Han and Zhao, Mingyue and Zhou, S Kevin},
  journal={npj Digital Medicine},
  volume={7},
  number={1},
  pages={286},
  year={2024},
  publisher={Nature Publishing Group UK London}
}

@inproceedings{jin2024debiased,
  title={Debiased noise editing on foundation models for fair medical image classification},
  author={Jin, Ruinan and Deng, Wenlong and Chen, Minghui and Li, Xiaoxiao},
  booktitle={International Conference on Medical Image Computing and Computer-Assisted Intervention},
  pages={164--174},
  year={2024},
  organization={Springer}
}

@inproceedings{xu2023fairadabn,
  title={Fairadabn: Mitigating unfairness with adaptive batch normalization and its application to dermatological disease classification},
  author={Xu, Zikang and Zhao, Shang and Quan, Quan and Yao, Qingsong and Zhou, S Kevin},
  booktitle={International Conference on Medical Image Computing and Computer-Assisted Intervention},
  pages={307--317},
  year={2023},
  organization={Springer}
}

@article{johnson2019mimic,
  title={MIMIC-CXR, a de-identified publicly available database of chest radiographs with free-text reports},
  author={Johnson, Alistair EW and Pollard, Tom J and Berkowitz, Seth J and Greenbaum, Nathaniel R and Lungren, Matthew P and Deng, Chih-ying and Mark, Roger G and Horng, Steven},
  journal={Scientific data},
  volume={6},
  number={1},
  pages={317},
  year={2019},
  publisher={Nature Publishing Group UK London}
}

@inproceedings{jackson2025enhancement,
  title={Enhancement of Fairness in AI for Chest X-ray Classification},
  author={Jackson, Nicholas J and Yan, Chao and Malin, Bradley A},
  booktitle={AMIA Annual Symposium Proceedings},
  volume={2024},
  pages={551},
  year={2025}
}

@article{hardt2016equality,
  title={Equality of opportunity in supervised learning},
  author={Hardt, Moritz and Price, Eric and Srebro, Nati},
  journal={Advances in neural information processing systems},
  volume={29},
  year={2016}
}

@inproceedings{zhang2022improving,
  title={Improving the fairness of chest x-ray classifiers},
  author={Zhang, Haoran and Dullerud, Natalie and Roth, Karsten and Oakden-Rayner, Lauren and Pfohl, Stephen and Ghassemi, Marzyeh},
  booktitle={Conference on health, inference, and learning},
  pages={204--233},
  year={2022},
  organization={PMLR}
}

@article{kim2024mdagents,
  title={Mdagents: An adaptive collaboration of llms for medical decision-making},
  author={Kim, Yubin and Park, Chanwoo and Jeong, Hyewon and Chan, Yik S and Xu, Xuhai and McDuff, Daniel and Lee, Hyeonhoon and Ghassemi, Marzyeh and Breazeal, Cynthia and Park, Hae W},
  journal={Advances in Neural Information Processing Systems},
  volume={37},
  pages={79410--79452},
  year={2024}
}

@article{sharma2024cxr,
  title={Cxr-agent: Vision-language models for chest x-ray interpretation with uncertainty aware radiology reporting},
  author={Sharma, Naman},
  journal={arXiv preprint arXiv:2407.08811},
  year={2024}
}

@article{glocker2023risk,
  title={Risk of bias in chest radiography deep learning foundation models},
  author={Glocker, Ben and Jones, Charles and Roschewitz, M{\'e}lanie and Winzeck, Stefan},
  journal={Radiology: Artificial Intelligence},
  volume={5},
  number={6},
  pages={e230060},
  year={2023},
  publisher={Radiological Society of North America}
}

@article{larrazabal2020gender,
  title={Gender imbalance in medical imaging datasets produces biased classifiers for computer-aided diagnosis},
  author={Larrazabal, Agostina J and Nieto, Nicol{\'a}s and Peterson, Victoria and Milone, Diego H and Ferrante, Enzo},
  journal={Proceedings of the National Academy of Sciences},
  volume={117},
  number={23},
  pages={12592--12594},
  year={2020},
  publisher={National Academy of Sciences}
}

@article{seyyed2021underdiagnosis,
  title={Underdiagnosis bias of artificial intelligence algorithms applied to chest radiographs in under-served patient populations},
  author={Seyyed-Kalantari, Laleh and Zhang, Haoran and McDermott, Matthew BA and Chen, Irene Y and Ghassemi, Marzyeh},
  journal={Nature medicine},
  volume={27},
  number={12},
  pages={2176--2182},
  year={2021},
  publisher={Nature Publishing Group US New York}
}

@article{zheng2023judging,
  title={Judging llm-as-a-judge with mt-bench and chatbot arena},
  author={Zheng, Lianmin and Chiang, Wei-Lin and Sheng, Ying and Zhuang, Siyuan and Wu, Zhanghao and Zhuang, Yonghao and Lin, Zi and Li, Zhuohan and Li, Dacheng and Xing, Eric and others},
  journal={Advances in neural information processing systems},
  volume={36},
  pages={46595--46623},
  year={2023}
}

@inproceedings{derner2025leveraging,
  title={Leveraging large language models to measure gender representation bias in gendered language corpora},
  author={Derner, Erik and De La Fuente, Sara Sansalvador and Guti{\'e}rrez, Yoan and Pozo, Paloma Moreda and Oliver, Nuria M},
  booktitle={Proceedings of the 6th Workshop on Gender Bias in Natural Language Processing (GeBNLP)},
  pages={468--483},
  year={2025}
}

@article{hanna2025assessing,
  title={Assessing racial and ethnic bias in text generation by large language models for health care--related tasks: Cross-sectional study},
  author={Hanna, John J and Wakene, Abdi D and Johnson, Andrew O and Lehmann, Christoph U and Medford, Richard J},
  journal={Journal of Medical Internet Research},
  volume={27},
  pages={e57257},
  year={2025},
  publisher={JMIR Publications Toronto, Canada}
}

@inproceedings{kim2024m,
  title={" I'm Not Sure, But...": Examining the Impact of Large Language Models' Uncertainty Expression on User Reliance and Trust},
  author={Kim, Sunnie SY and Liao, Q Vera and Vorvoreanu, Mihaela and Ballard, Stephanie and Vaughan, Jennifer Wortman},
  booktitle={Proceedings of the 2024 ACM conference on fairness, accountability, and transparency},
  pages={822--835},
  year={2024}
}

@article{jin2025see,
  title={See-in-Pairs: Reference Image-Guided Comparative Vision-Language Models for Medical Diagnosis},
  author={Jin, Ruinan and Huang, Gexin and Shen, Xinwei and Zhang, Qiong and Tan, Yan Shuo and Li, Xiaoxiao},
  journal={arXiv preprint arXiv:2506.18140},
  year={2025}
}
%




\end{document}